\documentclass[letterpaper, 10 pt, conference]{ieeeconf}  

\IEEEoverridecommandlockouts

\usepackage{cite} 
\usepackage{graphics} 
\usepackage{epsfig} 
\usepackage{mathptmx} 
\usepackage{times} 
\usepackage{amsmath} 
\usepackage{amssymb}  
\usepackage{multirow}
\usepackage{array}
\newcolumntype{P}[1]{>{\centering\arraybackslash}p{#1}}
\usepackage{algorithm}
\usepackage{algpseudocode}

\title{\LARGE \bf
3D-FFS: Faster 3D object detection with Focused Frustum Search in sensor fusion based networks
}

\author{Aniruddha Ganguly$^{1}$\textsuperscript{\textdagger}, Tasin Ishmam$^{1}$\textsuperscript{\textdagger}, Khandker Aftarul Islam$^{1}$\textsuperscript{\textdagger}, \\ Md Zahidur Rahman$^{2}$ and Md. Shamsuzzoha Bayzid$^{1}$
\thanks{$^{1}$Aniruddha Ganguly, Tasin Ishmam and Khandker Aftarul Islam and Md. Shamsuzzoha Bayzid are with Department of Computer Science and Engineering,
        Bangladesh University of Engineering and Technology, Dhaka, Bangladesh
        {\tt\small shams\_bayzid@cse.buet.ac.bd}}%

\thanks{$^{2}$Md Zahidur Rahman is with AWS at Amazon, Palo Alto, California
        {\tt\small zahidbuet106@gmail.com}}%

\thanks{\textsuperscript{\textdagger} These authors contributed equally to this work.}
}

\begin{document}

\maketitle
\thispagestyle{empty}
\pagestyle{empty}

\begin{abstract}

In this work we propose 3D-FFS, a novel approach to make sensor fusion based 3D object detection networks significantly faster using a class of computationally inexpensive heuristics. Existing sensor fusion based networks generate 3D region proposals by leveraging inferences from 2D object detectors. However, as images have no depth information, these networks rely on extracting semantic features of points from the entire scene to locate the object. 
By leveraging aggregated intrinsic properties (\textit{e.g.} point density) of point cloud data, 3D-FFS can substantially constrain the 3D search space and thereby significantly reduce training time, inference time and memory consumption without sacrificing accuracy. To demonstrate the efficacy of 3D-FFS, we have integrated it with Frustum ConvNet (F-ConvNet), a prominent sensor fusion based 3D object detection model. We assess the performance of 3D-FFS on the KITTI dataset.
Compared to F-ConvNet, we achieve improvements in training and inference times by up to 62.80\% and 58.96\%, respectively, while reducing the memory usage by up to 58.53\%. 
Additionally, we achieve 0.36\%, 0.59\% and 2.19\% improvements in accuracy for the Car, Pedestrian and Cyclist classes, respectively. 3D-FFS shows a lot of promise in domains with limited computing power, such as autonomous vehicles, drones and robotics where LiDAR-Camera based sensor fusion perception systems are widely used.

\end{abstract}

\section{Introduction}

In recent times, 3D object detection is gaining increased attention due to its wide applications in various areas such as autonomous vehicles, robotics and drones. The nature of these use-cases demand real-time response. As a result, ensuring reasonable performance under real world conditions is one of the most important research directions in recent times. Due to the inability of conventional cameras to accurately understand depth and perform localization in a scene, 3D sensors such as LiDAR are attracting considerable interest from both industry and academia. 3D sensor data (captured using LiDAR sensors or depth camera) are often represented in point clouds.

Point cloud data is known to be high-dimensional and sparse in nature. As a result, processing this data with neural networks has high computational overhead. Besides, storing each point cloud instance requires much more memory than conventional 2D data. Due to vast amounts of data that can be generated by various systems like autonomous vehicles and drones, training time is another major bottleneck for computationally demanding models. For object detection use-cases, points that belong to unrelated objects in the scene are considered to be noise and do not contain relevant information about the object of interest. As a result, identification and elimination of this noise is expected to speed up computation and remove a bottleneck in 3D object detection pipelines.

In this study, we show that by using computationally inexpensive heuristics that leverage the aggregated intrinsic properties of 3D data, such as point cloud density, it is possible to constrain the \textit{Region of Interest} (RoI) for the object's location in a scene. By integrating this information into 3D object detection networks, we can significantly reduce the search space and thus improve inference time, training time as well as memory consumption requirements of the model. Since this approach does not use any point-wise semantic information from the scene, insights can be drawn from the 3D data without any expensive computational steps.

Most existing works on 3D object detection can be broadly classified into two 
general approaches. One is grid or voxel based methods and the other is point based methods. In voxel based methods, the 3D point cloud is converted into 3D voxels \cite{Song2016deepslidingshapes, Chen_2019_ICCV_fast_pointrcnn, Zhou_2018_voxelnet, Maturana2015VoxNet, OctNet} and processed using 3D Convolutional Neural Networks (3D CNN). Additionally, some approaches convert the point cloud into 2D birds eye view maps which can leverage the existing 2D object detection models \cite{Yang_2018_Pixor, Chen_2017_MV3D}. Voxels are more compact representations compared to point clouds. However, these transformations may obscure natural invariances of the data and result in the loss of information about fine grained details. This loss of information degrades accuracy and localization performance. Therefore, in order to improve performance, different techniques for working directly with the point cloud data are gaining interest. Another issue with voxel based methods is that they use LiDAR as the primary sensor for semantic understanding and object detection. However, various industrial applications are already heavily invested in camera based sensor suites, where LiDAR is simply used as an auxiliary sensor.

Among various methods to detect objects in 3D, we focus on sensor fusion based methods due to their wide adoption in industries that use LiDAR as an auxiliary sensor to the existing suite of camera based perception systems. Some of the prominent networks using sensor fusion are frustum based, which generate a 3D frustum shaped RoI for each 2D object proposal. These approaches \cite{qi_frustum_2018, wang_frustum_2019, SIFRNet_2019, RoarNet_2019} rely on the PointNet \cite{qi2016pointnet, Pointnet++} architecture and directly extract features from raw point clouds for 3D object detection. Frustum based approaches have impressive performance and accuracy. However, as these methods need to compute semantic information per point in each frustum, processing frustums directly can be computationally expensive due to their volume. Performance may also be impacted by the fact that such sensor fusion based approaches generate 3D bounding box for each identified object by processing the entire point cloud space of a frustum. As a result, inference is often slow and unfit for practical use-cases. The most computationally expensive step of such frustum based methods is the 3D semantic information extraction phase where 3D point clouds are converted into higher dimensional feature vectors. 

Our contribution in this paper is to reduce this 3D search space using inexpensive, yet effective heuristics and thus improve training time, inference time, memory consumption and performance. We propose 3D-FFS (Faster \textbf{3D} object detection with \textbf{F}ocused \textbf{F}rustum \textbf{S}earch), which is a class of heuristics that uses the aggregated intrinsic properties of 3D data such as point density, point geometries, etc. to provide an approximate location of the object in the RoI. Finally, this information is leveraged to generate a constrained RoI.

Our proposed heuristic is generalized and can be used in tandem with various point cloud based methods. In order to demonstrate the efficacy of our approach, we have integrated this heuristic within the end-to-end pipeline of Frustum ConvNet (F-ConvNet) \cite{wang_frustum_2019}. F-ConvNet is a prominent sensor fusion based 3D object detection model that also adopts a frustum based approach. This model uses an end-to-end deep learning approach for 3D bounding box detection. F-ConvNet generates 3D region proposals based on bounding boxes from 2D object detectors. However, as 2D object detectors cannot infer the exact location of the object in 3D space, the model relies on extracting semantic information from the entire 3D RoI to detect the object of interest. As shown in Table-\ref{inference-time-table}, the inference time of the 3D-FFS-integrated F-ConvNet pipeline is lower than that of the original F-ConvNet model, making our end-to-end pipeline more suitable for latency-sensitive applications. By integrating 3D-FFS into the F-ConvNet pipeline, we demonstrate that it is possible to use aggregated intrinsic properties of the point cloud to effectively constrain the 3D region of interest and improve performance.

We find that we are able to reduce inference time by up to 58.96\% and training time by up to 62.80\% after integrating our heuristic into the F-ConvNet Pipeline. Furthermore, the model's memory usage is reduced by up to 58.53\%. Notably, this improves upon the original F-ConvNet model, albeit being substantially faster than F-ConvNet. Experimental results suggest that 3D-FFS improves the accuracy on Easy, Moderate, Hard categories in the KITTI Object Detection Benchmark by 0.14\%, 0.36\% and 0.10\%, respectively, for the Car class, by 0.39\%, 0.59\% and 7.32\%, respectively, for the Pedestrian class and by 2.09\%, 2.39\% and 9.43\%, respectively, for the Cyclist class, compared to the original F-ConvNet model. This significantly improves performance compared to the original F-ConvNet and makes our end-to-end pipeline much more suitable for real world use-cases.

To summarize, our contributions are as follows: 
\begin{itemize}
    \item We identify that certain aggregated intrinsic properties of the 3D point cloud can be used to constrain the search space and thus enhance the performance of 3D object detectors.
    
    \item We develop 3D-FFS, a lightweight and modular heuristic for search space reduction in sensor fusion based 3D object detection models. 
    
    \item We integrate 3D-FFS in the F-ConvNet pipeline and improve its accuracy, training time, inference time and memory consumption. 
\end{itemize}

\begin{figure*}[!h]
	\begin{center}
	\includegraphics[width=0.95\linewidth]{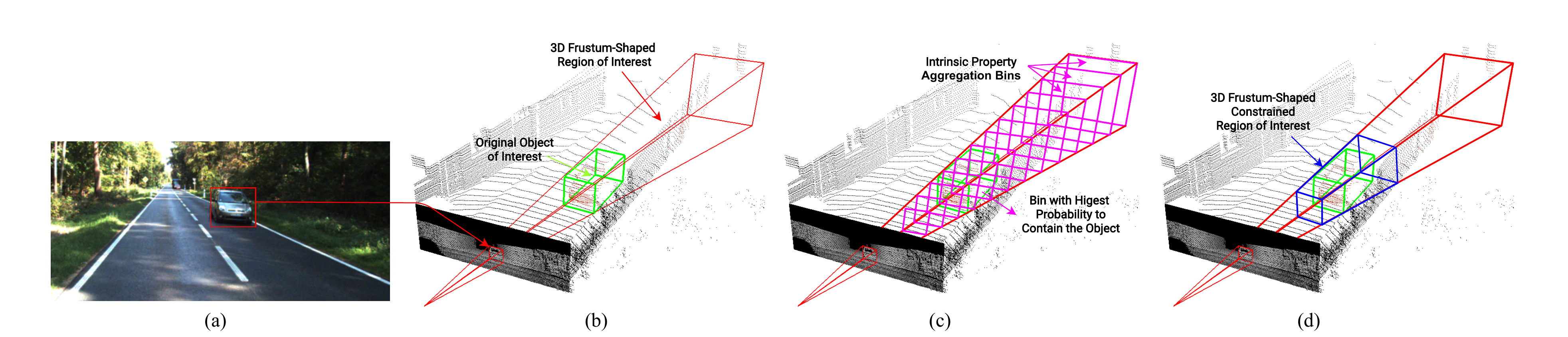}
	\caption[]{ Overview of 3D-FFS. (a) Generating 2D bounding box from RGB images for the object of interest. (b) Generating a 3D Region of Interest from a 2D bounding box. (c) Identification of the bin with the highest probability to contain the object using aggregated intrinsic properties. (d) Constraining the Region of Interest to a reduced length $h$. Part of the images used in this figure were obtained from \cite{FPointnets_2017}.}
	\label{Fig:heuristic}
	\end{center}
\end{figure*}

\section{Related Works} \label{related-work}

In this section we review existing frustum based 3D object detection models. We confine our discussion to models that are compatible with our problem domain and relevant to our approach.

A 3D object detector takes the point cloud of a scene as its input and produces an oriented 3D bounding box surrounding each detected object. Frustum based methods are some of the most prominent sensor fusion based 3D object detection methods. These methods leverage bounding boxes given by 2D object detectors to generate 3D frustum shaped region proposals. F-PointNets \cite{FPointnets_2017} is a pioneer among frustum based methods. It consists of three modules: frustum proposal module, 3D instance segmentation module and 3D amodal bounding box estimation module. The frustum proposal module proposes 3D frustums based on the bounding boxes generated by a 2D object detector. The 3D instance segmentation module then applies PointNet on the points collected from the entire proposed frustum to extract features. The 3D amodal bounding box estimation module uses these extracted features to infer the coordinates of the 3D bounding box of the object. In the 3D instance segmentation module the semantic information is collected from the entire proposed frustum which slows down the process and significantly increases memory consumption. It also introduces more noise in the extracted features, thereby reducing accuracy. Zhao \textit{et al.} \cite{SIFRNet_2019} proposed a new network architecture, Scale Invariant Feature Reweighting Network (SIFRNet), which uses the front view images and frustum shaped point clouds for 3D object detection. This architecture consists of multiple modules. The Point-UNet module takes 3D points in the entire frustum as the input and for each point outputs the probability of it being a part of the object of interest. This module also uses information from the 2D image. Next, a T-Net uses the points corresponding to the object of interest to estimate the center of that object. Finally, the Point-SENet (Point Squeeze-and-Excitation Network) module, which is inspired by the Squeeze and Excitation Networks proposed by Hu \textit{et al.} \cite{SE_Hu_2018}, is used to estimate the coordinates and orientation of the 3D bounding box. We notice that the entire frustum point cloud is used as the input in the Point-UNet module which slows down the model drastically. As we are detecting the probability of the points being part of the object of interest, a substantial number of points can be discarded just by analyzing their positions. Wang \textit{et al.} \cite{wang_frustum_2019} have proposed a novel method, Frustum ConvNet (F-ConvNet). F-ConvNet is a prominent 3D object detection model. F-ConvNet generates a sequence of frustums for each 2D region. Semantic information is extracted and aggregated as frustum level features from the point cloud by applying PointNet on the entire sequence of frustums. These frustum level features are transformed into a 2D feature map, which is then fed into a Fully Convolutional Network (FCN) for 3D box estimation. F-ConvNet is slowed down significantly by the semantic information extraction step as this needs to be done for the entire sequence of frustums. Evidently, one of the major bottlenecks of these approaches is the computational overhead of processing large point clouds. However, as we will show in this study, using region proposals from a 2D object detector and aggregated intrinsic properties of 3D data, the length of the frustum can be constrained without losing information. Hence, our proposed heuristic 3D-FFS is designed to reduce the computational overhead of processing large point clouds.

\section{Methods} \label{methodology}
 In this section we explain the methodology used by 3D-FFS for constraining the RoI in sensor fusion based object detection models. In Sec. \ref{review-fconv} we provide a short overview of the F-ConvNet model in order to provide insights into the pipeline we will work with. Next, in Sec. \ref{methodology-heuristic}, we explain our own contribution and how intrinsic aggregate properties of 3D data can be used for search space reduction. 

 \subsection{Review of the F-ConvNet model} \label{review-fconv}
 F-ConvNet is a point cloud based object detection model that supports end-to-end learning and prediction of amodal 3D bounding boxes. A major underlying assumption of the F-ConvNet model is the availability of calibrated RGB images. RGB images are used to generate 2D bounding boxes for each object of interest using state-of-the-art object detectors \cite{RRC, MSCNN, conf/eccv/LiuAESRFB16SSD, NIPS2015fasterrcnn, Redmon2018YOLOv3AI}.
 Mapping the 2D region proposal to 3D space, this model generates a series of $T$ overlapping frustums by sliding a pair of parallel planes along the frustum axis\footnote{For each 2D object bounding box, a truncated pyramid or ``frustum'' can be formed in 3D space where the near plane is the 2D bounding box itself and the far plane extends outwards, possibly up to infinity. The axis perpendicular to the image plane and originating from the center of the 2D bounding box, is known as the frustum axis.} with equal stride $s$. In each frustum, groups of discrete unordered points are connected together. These points may belong to the object of interest as well as foreground or background objects.
 
 A single three-layer fully connected PointNet \cite{qi2016pointnet} is used to learn point-wise features from each of the $T$ Frustums. After iterating through each of the $T$ frustums, the PointNet model generates $T$ feature vectors $\{ \mathbf{f}_i \}_{i=1}^L$, with $\mathbf{f}_i \in \mathbb{R}^d$, one for each of the frustums processed. These $L$ vectors are stacked together to form a 2D feature map $\mathbf{F}$ of  size $L\times d$. This feature map is the input of the Fully Convolutional Network (FCN), which uses convolutions followed by de-convolutions to fuse features across frustums. Finally, the learned features are passed to the two detection headers (Classification and Regression) which predict the object class and bounding box (center co-ordinates, length, width, height and yaw angle), respectively. There is a variant of the F-ConvNet model which generates frustums at multiple resolutions instead of just one to provide a modest improvement in performance. We have used this variant in our experiments.
 
\subsection{Constraining search space using 3D-FFS} \label{methodology-heuristic}

In this section we present our proposed heuristic for constraining the search space of sensor fusion based 3D object detection models. The end-to-end pipeline of 3D-FFS is shown in Fig. \ref{Fig:heuristic}.

\subsubsection{Using RGB detectors to identify 3D frustum shaped Region of Interest (RoI)}

One of the underlying assumptions of our architecture is the availability of 2D images. These 2D images are leveraged to generate 3D RoIs using bounding boxes from state-of-the-art 2D object detectors. This 3D RoI is frustum shaped where the near plane is the 2D bounding box itself and the far plane extends outwards, possibly up to infinity. For practical purposes, we constrain the far plane of a frustum based on the hardware limits of the LiDAR sensor. In case of the KITTI dataset, the far plane is constrained to a distance of 70 meters from the near plane. Assuming that the RoI extrapolated from the 2D bounding box is accurate, this frustum shaped region is aware of object boundaries and accurately encloses the object.

\subsubsection{Analysis of aggregated intrinsic properties of the point cloud in the Region of Interest}

Each frustum has only one object of interest which we can approximately locate by using various aggregated intrinsic properties of the point cloud. We divide the frustum shaped RoI into granular bins by iterating along the frustum axis with a stride of $bin\_length$. By iterating across the bins, we calculate point cloud densities in each bin. Finally, the bin with the highest point density is identified, which is expected to contain the object of interest. We denote by $c$ the distance of this bin's center from the origin of the frustum axis.

In order to be resilient against foreground and background noise such as small stray objects, we allow the points of a certain bin to contribute a weight $w$ to its neighboring bins. For a particular point, we allow $neighbor\_bins$ bins on either side of its parent bin to receive weight contributions from the point. As a result, a dense but irrelevant point cluster is less likely to influence our results, making this approach robust against noise.

\subsubsection{Constraining the Region of Interest}

In the previous section we have discussed how we identify the bin with highest point density and how we calculated $c$, which is the distance of its center from the origin of the frustum axis. Next, we constrain the original RoI along the frustum axis to a reduced length of $h$. The near plane of the constrained RoI is at a distance $c-\frac{h}{2}$ from the origin of the frustum axis while the far plane is at a distance $c+\frac{h}{2}$. Our experimental results (Sec. \ref{experiments-integration-fconvnet}) suggest that $h$ can be reduced to around one-third of the length of the original RoI, while improving upon the accuracy of F-ConvNet. We also significantly improve training time, inference time and memory consumption compared to the original model.

\begin{table*}[h]
        \vspace{0.20cm}

		\caption{ \label{Tab:KITTI_TEST_3D} 3D object detection AP (\%) on KITTI Validation set}
	\begin{center}
		\begin{tabular}{c|ccc|ccc|ccc}
			\hline
			\multirow{2}{*}{Constrained RoI Length (meter)}         & \multicolumn{3}{c|}{Cars} & \multicolumn{3}{c|}{Pedestrians} & \multicolumn{3}{c}{Cyclists}                                         \\ \cline{2-10}
		    & Easy & Moderate & Hard & Easy & Moderate & Hard & Easy & Moderate & Hard \\ \hline
			10 & 85.88 & 83.29 & 75.84 & 63.24 & 59.25 & 58.25 & 88.87 & 86.28 & 79.27 \\
			20 & 87.56 & 85.43 & 78.24 & \textbf{69.36} & \textbf{62.49} & \textbf{62.26} & 89.32 & 87.32 & 82.66 \\
			26 & \textbf{88.39} & 85.78 & 78.43 & 68.52 & 61.32 & 59.93 & \textbf{90.53} & \textbf{88.64} & \textbf{87.72} \\
			30 & 88.22 & \textbf{86.14} & \textbf{78.52} & 67.95 & 61.49 & 60.00 & 89.63 & 88.14 & 84.02 \\
			40 & 87.54 & 85.46 & 78.20 & 68.98 & 62.28 & 59.10 & 89.91 & 87.57 & 82.83 \\
			50 & 87.95 & 85.60 & 78.32 & 68.79 & 62.24 & 61.92 & 89.58 & 87.92 & 83.99 \\
			60 & 88.36 & 85.67 & 78.39 & 68.12 & 62.12 & 58.10 & 89.73 & 88.26 & 85.62 \\
            70 & 87.99 & 85.65 & 78.32 & 68.08 & 61.85 & 60.17 & 88.50 & 87.29 & 84.19 \\
            \textit{F-ConvNet} & 88.25 & 85.78 & 78.42 & 68.97 & 61.90 & 54.94 & 88.44 & 86.25 & 78.29 \\ \hline
		\end{tabular}
	\end{center}
\end{table*}

\begin{table*}[h]
		\caption{\label{Tab:KITTI_VALIDATION_BEV} 3D object localization AP (BEV) (\%) on KITTI Validation set}
	\begin{center}
		\begin{tabular}{c|ccc|ccc|ccc}
			\hline
			\multirow{2}{*}{Constrained RoI Length (meter)} & \multicolumn{3}{c|}{Cars} & \multicolumn{3}{c|}{Pedestrians} & \multicolumn{3}{c}{Cyclists} 
			\\ \cline{2-10}
			& Easy & Moderate & Hard & Easy & Moderate & Hard & Easy & Moderate & Hard \\ \hline
			10 & 96.58 & 88.18 & 86.54 & 70.57 & 67.11 & 61.88 & 95.01 & 87.76 & 85.63 \\
			20 & 96.92 & 89.45 & 88.97 & \textbf{73.16} & \textbf{72.59} & \textbf{66.83} & 93.45 & 89.40 & 87.29 \\
			26 & 95.85 & 89.36 & 88.91 & 71.78 & 70.25 & 64.89 & \textbf{97.02} & \textbf{89.58} & \textbf{87.47} \\
			30 & 96.00 & 89.43 & 88.85 & 71.58 & 71.81 & 65.23 & 96.76 & 89.28 & 87.23 \\
			40 & 96.06 & 89.56 & 88.99 & 72.31 & 67.76 & 65.87 & 91.46 & 88.77 & 86.61 \\
			50 & 97.88 & 89.48 & \textbf{89.22} & 72.89 & 71.11 & 66.69 & 93.50 & 89.27 & 85.59 \\
			60 & \textbf{97.98} & 89.68 & 89.01 & 72.17 & 69.18 & 65.07 & 96.92 & 89.33 & 87.18 \\
			70 & 97.76 & 89.56 & 88.99 & 71.82 & 67.49 & 65.24 & 91.40 & 88.88 & 85.24 \\
			\textit{F-ConvNet} & 97.84 & \textbf{89.70} & 89.05 & 72.45 & 64.77 & 64.06 & 89.27 & 87.92 & 79.45 \\ \hline
		\end{tabular}
	\end{center}
\end{table*}

\begin{figure*}
	\begin{center}
	\includegraphics[width=0.80\linewidth]{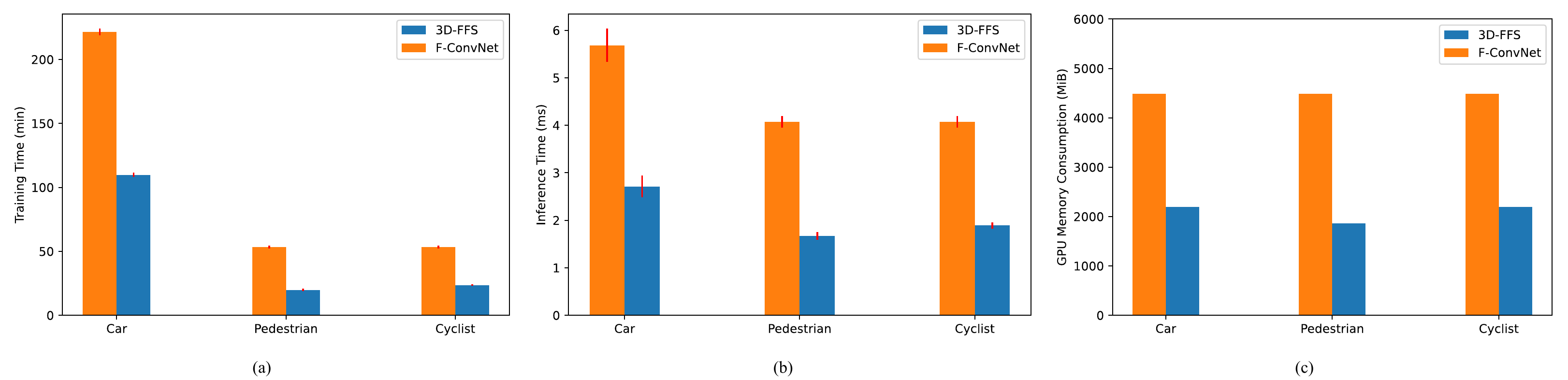}
	\includegraphics[width=0.80\linewidth]{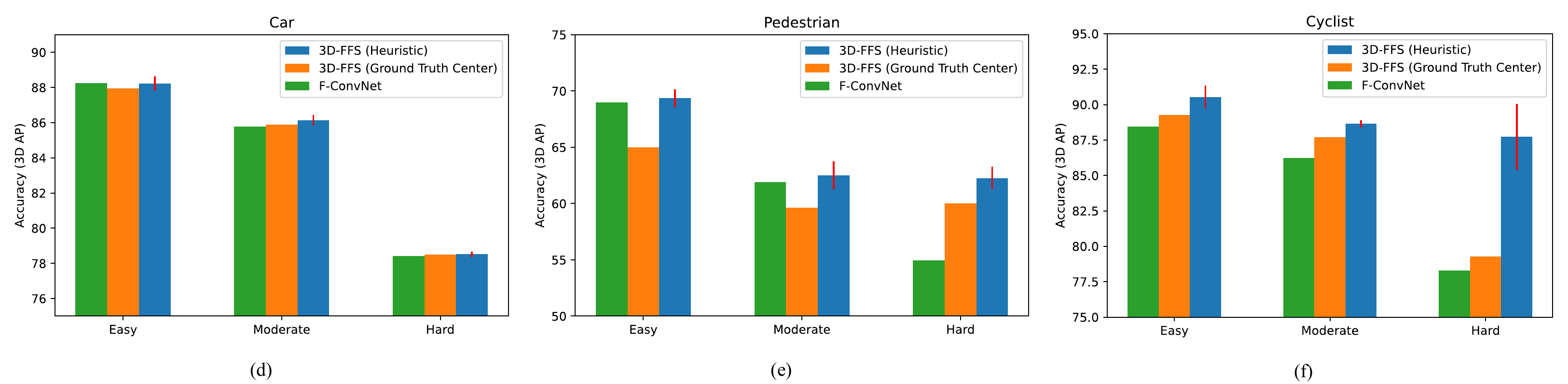}
	\hspace*{0.30cm}\includegraphics[width=0.82\linewidth]{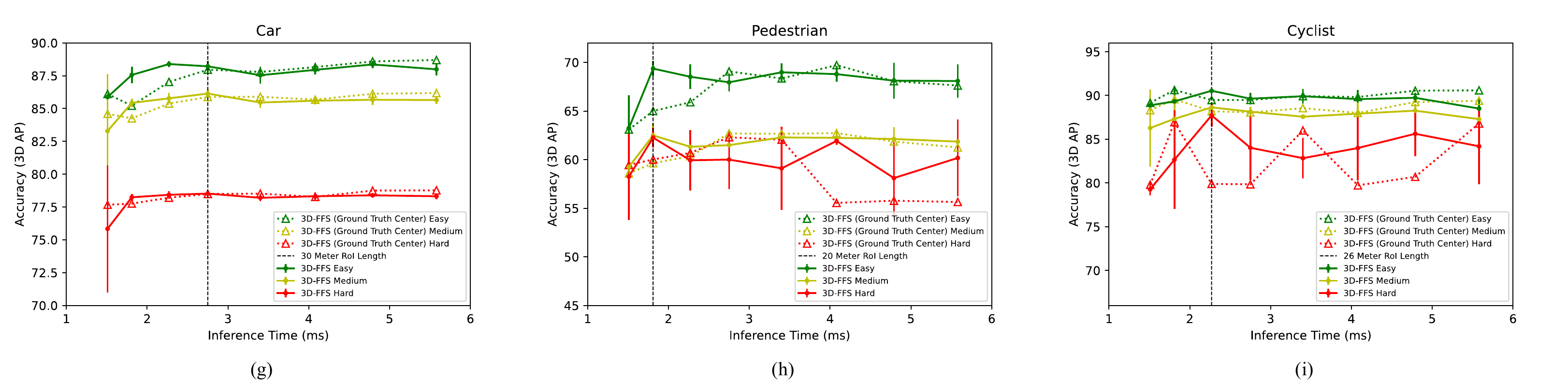}
	\caption[]{Comparison of 3D-FFS enhanced F-ConvNet and original F-ConvNet on the KITTI training and validation set. The average results over five independent runs of the experiments are shown. (a) Training time comparison with original F-ConvNet, (b) Inference time comparison with original F-ConvNet, (c) Memory Consumption during training, (d) - (f) 3D Average precision on the KITTI validation set, (g) - (i) Comparison of 3D average precision and inference time for 10m-70m constrained RoI lengths.
	For the figures (a) - (f), experiments for the Car, Pedestrian and Cyclist classes were conducted with 30 meter, 20 meter and 26 meter constrained RoI lengths respectively.}
	\label{Fig:heuristic-charts}
	\end{center}
\end{figure*}

\section{Results and Discussion}\label{xperiments}
We conduct experiments on the KITTI Dataset \cite{kittidataset}, containing 7,481 training images, 7,518 test images and their corresponding point clouds. We report our findings by generating a validation set similar to \cite{Chen_2017_MV3D}. We conduct our experiments using the official KITTI evaluation protocol \cite{kitti_evaluation_are_we_ready..}, by setting 0.7 IoU threshold for the Car class and 0.5 for the Pedestrian and Cyclist classes. 

\subsection{Evaluation of our heuristic} \label{experiments-heuristic-eval}
To evaluate our proposed heuristic, we measure how much the center of the constrained RoI generated by our heuristic differs from the ground truth center of the object of interest. We calculate the Root Mean Squared Error (RMSE) between the ground truth center of the object and the center of the constrained RoI proposed by our heuristic. For the Car class, we obtain an RMSE of 4.98m and for the Pedestrian and Cyclist classes, we obtain an RMSE of 5.43m. These values are consistent with our experiments. We observe that our heuristic performs best when the length of the constrained RoI is within the 20m - 30m range. The average length of objects belonging to the Car class in the KITTI Dataset is 3.88m. An RMSE of around 5 meters is acceptable since the RoI is expected to contain the object of interest.

\subsection{Effect of integration of our heuristic with F-ConvNet} \label{experiments-integration-fconvnet}

To better understand the effect of our heuristic on F-ConvNet, we train the 3D-FFS integrated F-ConvNet pipeline with various configurations and report our findings.

\subsubsection{Feeding Ground Truth Center} \label{experiments-feeding-ground-truth}

This series of experiments establishes a baseline for a heuristic that leverages object center depth as the intrinsic property which is used to constrain the RoI. Using the ground truth centers of the object of interest we constrain the original RoI to various smaller lengths. We determine the accuracy of the F-ConvNet model under these conditions. This helps us compare the performance of a depth based heuristic with that of a point density based heuristic. The results are shown in Fig. \ref{Fig:heuristic-charts} (d-i). Interestingly -- as we can observe from these results -- using our density based heuristic usually gives us better accuracies across most categories of the KITTI dataset. Future studies need to investigate this further to arrive at a more concrete conclusion regarding the best heuristic choice in a dataset and model agnostic setting.

\begin{table}[!h]
    \vspace{0.20cm}
    \caption[Memory Usage, Inference Time (Infer) and Training Time (Train) for KITTI Dataset]{\label{inference-time-table} Memory Usage, Inference Time (Infer) and Training Time (Train) for KITTI Dataset\footnotemark}
    \begin{center}
        \begin{tabular}
            {|P{1.42cm}|P{0.90cm}|P{0.7cm}|P{0.7cm}|P{1.0cm}|P{0.90cm}|}
            \hline
            \multirow{2}{1.42cm}{\centering \textbf{Constrained RoI Length (meter)}} & \multirow{2}{0.90cm}{\centering \textbf{Memory Usage (MiB)}} & \multicolumn{2}{c|}{\textbf{Car}}              & \multicolumn{2}{c|}{\textbf{Pedestrian \& Cyclist}} 
            \\ \cline{3-6} 
            &&
            \textbf{Infer (ms)} & \textbf{Train (min)} & \textbf{Infer (ms)} & \textbf{Train (min)} \\ \hline
                            10 & 1331 & 1.29 &  54.52 & 1.21 & 13.04 \\ \hline
                            20 & 1859 & 1.98 &  82.00 & 1.67 & 19.86 \\ \hline
                            26 & 2197 & 2.36 &  98.88 & 1.89 & 23.56 \\ \hline
                            30 & 2405 & 2.71 & 109.83 & 2.13 & 25.92 \\ \hline
                            40 & 2925 & 3.30 & 140.96 & 2.51 & 33.06 \\ \hline
                            50 & 3443 & 3.96 & 164.72 & 2.97 & 39.43 \\ \hline
                            60 & 3823 & 4.54 & 191.52 & 3.46 & 46.17 \\ \hline
                            70 & 4483 & 5.62 & 223.01 & 3.92 & 53.49 \\ \hline
            \textit{F-ConvNet} & 4483 & 5.68 & 221.58 & 4.07 & 53.39 \\ \hline
        \end{tabular}
    \end{center}
\end{table}

\subsubsection{Comparing performance across Regions of Interest of various lengths} \label{experiments-estimating-depth-point-cloud}
In these experiments, we run the 3D-FFS integrated F-ConvNet pipeline to generate a discrete set of constrained Regions of Interest (RoI) of lengths 10, 20, 26, 30, 40, 50, 60 and 70 meters. We have observed that the choice of RoI length affects our training time, inference time and accuracy. The choice of RoI length can be governed by the target class of the object of interest. So there is scope to optimize RoI length based on the standard dimension of the target class proposed by the 2D detector. For generality, we primarily focus on the car class. The parameters for our heuristic, i.e. $bin\_length$ and $neighbor\_bins$ are set to 0.75m and 7, respectively. We settle on these values after an exhaustive grid search using $bin\_length$s in the range 0m to 2m at 0.05m steps and $neighbor\_bins$ in the range 0 to 10. We observe based on our experimental results (not shown due to space constraints) that the 3D-FFS integrated pipeline is not sensitive to the hyper parameters $w$, $neighbor\_bins$ and $bin\_length$ with respect to detection accuracy, training time and inference time.

Table \ref{Tab:KITTI_TEST_3D} and \ref{Tab:KITTI_VALIDATION_BEV} demonstrate the accuracy of 3D-FFS in comparison with the original F-ConvNet. The most important observation is the ability of 3D-FFS to drastically reduce training time, inference time and memory consumption while also increasing the accuracy. Notably, 3D-FFS matches the performance of F-ConvNet even with a constrained RoI length of around 10 meters, whereas F-ConvNet uses a 70 meter RoI length (See Tables \ref{Tab:KITTI_TEST_3D} and \ref{Tab:KITTI_VALIDATION_BEV}). Also, we observe from Table-\ref{inference-time-table} that, by using a constrained RoI length of 30m, inference time reduces by 52.28\% for the Car class and using an RoI length of 26m, it reduces by 53.56\% for the Cyclist class, while giving us better accuracy than the original F-ConvNet. Additionally, we obtain a 58.96\% reduction in inference time for the Pedestrian class with a slight increase in accuracy, using a constrained RoI length of 20m. Training time is also reduced by 50.43\%, 62.80\% and 55.87\% for the car, pedestrian and cyclist classes, respectively. Correspondingly, we observe that the GPU memory usage reduces by 58.53\%, 50.99\% and 46.35\% for constrained RoI lengths of 20m, 26m and 30m respectively. A short accompanying video is provided with this paper to summarize our methodology and experimental results.

\section{Conclusion}

In this work we have presented 3D-FFS, a generic approach to faster yet accurate 3D object detection in sensor fusion based methods. By constraining the region of interest using computationally inexpensive heuristics based on aggregated intrinsic properties of the point cloud, 3D-FFS significantly improves memory consumption, training time and inference time.
We have demonstrated the efficacy of our proposed approach by integrating it with the F-ConvNet pipeline. Experimental results show that our proposed method matches or improves upon the accuracy of the original F-ConvNet, albeit being significantly faster. The timing of this approach seems appropriate since accurate 3D object detection is crucial and becoming popular for various latency-sensitive applications such as autonomous vehicles, drones and delivery robots. Future studies need to investigate its application to other point cloud based networks, such as Frustum PointNets, SIFRNet, RoarNet etc. Additionally, future studies need to investigate the utilization of other 3D point cloud properties such as point geometry, object depth etc. to enhance the detection of object orientation and location in 3D.

\footnotetext{Experiments were run on an Ubuntu 18.04 machine with an AMD Ryzen 7 3700X CPU running at 3.6 GHz using 48 GB RAM and NVIDIA RTX 2060 Super GPU.}

\addtolength{\textheight}{-12cm}   






\bibliographystyle{IEEEtran}
\bibliography{root}

\end{document}